\documentclass{article}

\usepackage{arxiv}

\usepackage[utf8]{inputenc} 
\usepackage[T1]{fontenc}    
\usepackage{hyperref}       
\usepackage{url}            
\usepackage{booktabs}       
\usepackage{amsfonts}       
\usepackage{nicefrac}       
\usepackage{microtype}      
\usepackage{lipsum}		
\usepackage{graphicx}
\usepackage{natbib}
\usepackage{doi}
\usepackage{amsmath}
\usepackage{amssymb}
\usepackage{mathtools}
\usepackage{subcaption}
\usepackage{amsthm}
\usepackage{enumitem}
\usepackage{wrapfig}
\usepackage{multirow}
\usepackage{pifont}

\usepackage{tikz}
\usetikzlibrary{calc}
\usetikzlibrary{shapes,decorations,arrows,calc,arrows.meta,fit,positioning}
\tikzset{
    -Latex,auto,node distance = 25pt and 25pt,semithick,
    state/.style ={circle, draw, minimum width = 0.7 cm},
    point/.style = {circle, draw, inner sep=0.04cm,fill,node contents={}},
    bidirected/.style={Latex-Latex,dashed},
    el/.style = {inner sep=2pt, align=left, sloped}
}
\tikzset{
    smallnode/.style={state,draw,inner sep=2,font=\scriptsize}
}
\tikzset{
    smallbluenode/.style={state,draw,blue,inner sep=2,font=\scriptsize}
}

\theoremstyle{plain}

\theoremstyle{definition}

\newtheorem{assumption}{Assumption}
\theoremstyle{remark}

\newcommand\independent{\protect\mathpalette{\protect\independenT}{\perp}}
\def\independenT#1#2{\mathrel{\rlap{$#1#2$}\mkern2mu{#1#2}}}

\title{Deep Causal Inference for Point-referenced Spatial Data with Continuous Treatments}

\author{ 
{Ziyang Jiang} \\
Department of Civil and \\ Environmental Engineering \\
Duke University \\
\texttt{ziyang.jiang@duke.edu} \\
\And
{Zach Calhoun} \\
Department of Civil and \\ Environmental Engineering \\
Duke University \\
\texttt{zachary.calhoun@duke.edu} \\
\And
{Yiling Liu} \\
Department of Computational \\ 
Biology and Bioinformatics \\
Duke University \\
\texttt{yiling.liu@duke.edu} 
\AND
{Lei Duan} \\
Department of Civil and \\ Environmental Engineering \\
Duke University \\
\texttt{lei.duan@duke.edu}
\And
{David Carlson} \\
Department of Civil and \\ Environmental Engineering \\
Duke University\\
	\texttt{david.carlson@duke.edu} \\
}

\renewcommand{\shorttitle}{\textit{arXiv} Template}

\begin{document}
\maketitle

\begin{abstract}
Causal reasoning is often challenging with spatial data, particularly when handling high-dimensional inputs. To address this, we propose a neural network (NN) based framework integrated with an approximate Gaussian process to manage spatial interference and unobserved confounding. Additionally, we adopt a generalized propensity-score-based approach to address partially observed outcomes when estimating causal effects with continuous treatments. We evaluate our framework using synthetic, semi-synthetic, and real-world data inferred from satellite imagery. \footnotemark{} Our results demonstrate that NN-based models significantly outperform linear spatial regression models in estimating causal effects. Furthermore, in real-world case studies, NN-based models offer more reasonable predictions of causal effects, facilitating decision-making in relevant applications.
\end{abstract}

\footnotetext[1]{Code is available at: \url{https://github.com/jzy95310/deep_sci}.}


\section{Introduction}
\label{sec:1}

Drawing causal relationships from data is essential across multiple disciplines, such as economic studies \cite{howard2010supported}, drug discovery \cite{michoel2023causal}, and neuroscience experiments \cite{jiang2023causal}. However, many of the existing causal inference methodologies rely on a number of assumptions, including the Stable Unit Treatment Value Assumption (SUTVA) \cite{rosenbaum1983central,rubin2005causal} and the principle of strong ignorability (i.e., the absence of unmeasured confounders) \cite{pearl2009causal}. These assumptions can be violated when dealing with spatial data, which plays a crucial role in various scientific fields including environmental science \cite{haining2003spatial,banerjee2003hierarchical,cressie2015statistics,zheng2021local,jiang2022improving}, urban planning \cite{sudhira2003urban,scott2008role,schabenberger2017statistical,yan2022influencing,hu2023artificial}, agriculture \cite{rajesh2011application,plant2018spatial}, distributed networking \cite{zhu2022optimizing}, public health \cite{rushton2003public,waller2004applied,zhu2021twitter}, and beyond. In this paper, we focus on causal inference related to a specific type of spatial data: point-referenced data, where we observe realizations or measurements of an underlying spatial process at a fixed set of locations. For instance, a measurement might represent an observation of pollutant levels. While it is theoretically conceivable to assume the existence of pollutant levels across an entire spatial domain, in practice, we are limited to finite measurements obtained from ground monitoring stations.

There are \textcolor{black}{two} major differences between causal inference on non-spatial and spatial data \cite{reich2021review}. \textcolor{black}{First}, the treatment assigned to one unit may affect the outcomes of other units (usually adjacent or neighboring units) in the spatial domain, which we call \emph{spatial interference} or \emph{spillover effects}. \textcolor{black}{Also,} it is crucial to consider unobserved spatial confounding variables, as comprehensive data on all relevant confounders (such as climatological or regulatory factors) are almost never observed but are anticipated to exhibit spatial variation and affect both treatment and target variables in most scenarios \cite{papadogeorgou2019adjusting}. Moreover, when the scale of unobserved spatial confounding is greater than the spatial variation of the observed variables, including this effect reduces the bias of estimates of treatment effects \cite{paciorek2010importance}.

Significant progress has been made in causal inference for point-referenced spatial data in linear models. Specifically, to account for the unobserved spatial confounder, some common approaches include \textcolor{black}{introducing a spatial proximity term into the propensity score matching procedure \cite{papadogeorgou2019adjusting}, comparing observations that are spatially close but differ in treatment to eliminate the effect of the unobserved confounder \cite{imbens2008regression,bor2014regression,keele2015geographic}, or explicitly modeling the unobserved spatial confounder as a zero-mean Gaussian process (GP) \cite{marques2022mitigating,calhoun2024estimating}.} In addition, when dealing with spatial interference, researchers make different assumptions about the form of interference (e.g., group-based interference \cite{perez2014assessing,zigler2021bipartite}, network-based interference \cite{aronow2017estimating,tchetgen2021auto,ogburn2022causal}, process-based interference \cite{cross2019confronting}, etc.) and come up with different solutions such as structural equation model \cite{bovendorp2019defaunation} and Bayesian statistical model \cite{papadogeorgou2023spatial}.

\textcolor{black}{Nevertheless, these methods are not sufficiently effective as the causal effects of both unobserved confounders and spatial interference can be highly nonlinear. For example, when treatments, confounders, and targets are all inferred from satellite imagery, the relationship among these variables can be complex and heavily dependent on their spatial variation. 
Also, many of the existing works assume there is only one \emph{binary} treatment, which limits their applicability on spatial data, as we may wish to consider multiple, continuous treatments that commonly exist in environmental problems (e.g., variable amounts of tree cover, continuous levels of albedo, air pollution levels, etc.).} 

In light of these observations, in this study, we propose to integrate deep neural networks (NNs) with the spatial regression model for causal inference on point-referenced data. \textcolor{black}{We posit that NN-based spatial regression model would be able to capture nonlinear causal relationships in the presence of spatial interference and unobserved confounders.} Our contributions can be summarized as follows.

\begin{itemize} [leftmargin=*]
\item We explore spatial causal inference on point-referenced data under the potential outcomes framework with continuous treatments where the SUTVA assumption can be relaxed.
\item We develop a deep-neural-network-based spatial causal inference framework and integrate it with an \textcolor{black}{approximate} GP to handle treatments and measured confounders inferred from high-dimensional data.
\item We employ a generalized propensity-score-based approach to address the challenge of partially observed outcomes in the context of continuous treatments.
\item To validate the efficacy of our proposed methodology, we perform experiments on synthetic and semi-synthetic datasets, and real-world case studies that include satellite imagery.
\item We investigate various architectures of NNs and evaluate their impact on the performance of the proposed framework.
\end{itemize}


\section{Causal Inference for Point-referenced Spatial Data}
\label{sec:2}

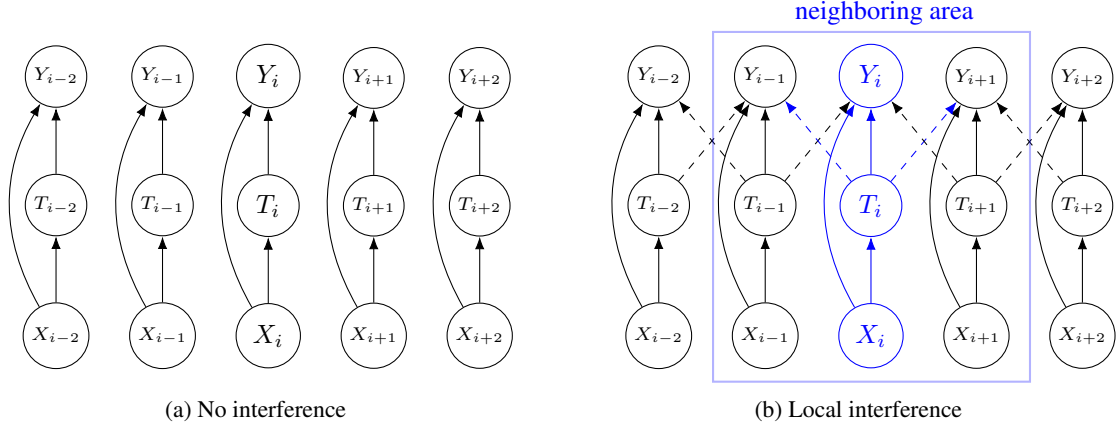
\begin{figure*}[t!]
\begin{subfigure}[t]{0.48\textwidth}
    \centering
    \begin{tikzpicture}
    \node[smallnode] (1) {$X_{i-2}$};
    \node[smallnode] (2) [above =of 1] {$T_{i-2}$};
    \node[smallnode] (3) [above =of 2] {$Y_{i-2}$};
    \node[smallnode] (4) [right =of 1, xshift=-10pt] {$X_{i-1}$};
    \node[smallnode] (5) [above =of 4] {$T_{i-1}$};
    \node[smallnode] (6) [above =of 5] {$Y_{i-1}$};
    \node[state,draw,inner sep=3.6] (7) [right =of 4, xshift=-10pt] {$X_i$};
    \node[state,draw,inner sep=3.9] (8) [above =of 7] {$T_i$};
    \node[state,draw,inner sep=4.1] (9) [above =of 8] {$Y_i$};
    \node[smallnode] (10) [right =of 7, xshift=-10pt] {$X_{i+1}$};
    \node[smallnode] (11) [above =of 10] {$T_{i+1}$};
    \node[smallnode] (12) [above =of 11] {$Y_{i+1}$};
    \node[smallnode] (13) [right =of 10, xshift=-10pt] {$X_{i+2}$};
    \node[smallnode] (14) [above =of 13] {$T_{i+2}$};
    \node[smallnode] (15) [above =of 14] {$Y_{i+2}$};
    \path (1) edge (2);
    \path (2) edge (3);
    \path (1) edge [bend left] (3);
    \path (4) edge (5);
    \path (5) edge (6);
    \path (4) edge [bend left] (6);
    \path (7) edge (8);
    \path (8) edge (9);
    \path (7) edge [bend left] (9);
    \path (10) edge (11);
    \path (11) edge (12);
    \path (10) edge [bend left] (12);
    \path (13) edge (14);
    \path (14) edge (15);
    \path (13) edge [bend left] (15);
    \draw[blue,thick,opacity=0.] ($(6.north west)+(-0.4,0.3)$)  rectangle ($(10.south east)+(0.4,-0.3)$);
    \end{tikzpicture}
    \caption{No interference}
    \label{fig:1a}
\end{subfigure}
\begin{subfigure}[t]{0.48\textwidth}
    \centering
    \begin{tikzpicture}
    \node[smallnode] (1) {$X_{i-2}$};
    \node[smallnode] (2) [above =of 1] {$T_{i-2}$};
    \node[smallnode] (3) [above =of 2] {$Y_{i-2}$};
    \node[smallnode] (4) [right =of 1, xshift=-10pt] {$X_{i-1}$};
    \node[smallnode] (5) [above =of 4] {$T_{i-1}$};
    \node[smallnode] (6) [above =of 5] {$Y_{i-1}$};
    \node[state,draw,blue,inner sep=3.6] (7) [right =of 4, xshift=-10pt] {$X_i$};
    \node[state,draw,blue,inner sep=3.9] (8) [above =of 7] {$T_i$};
    \node[state,draw,blue,inner sep=4.1] (9) [above =of 8] {$Y_i$};
    \node[smallnode] (10) [right =of 7, xshift=-10pt] {$X_{i+1}$};
    \node[smallnode] (11) [above =of 10] {$T_{i+1}$};
    \node[smallnode] (12) [above =of 11] {$Y_{i+1}$};
    \node[smallnode] (13) [right =of 10, xshift=-10pt] {$X_{i+2}$};
    \node[smallnode] (14) [above =of 13] {$T_{i+2}$};
    \node[smallnode] (15) [above =of 14] {$Y_{i+2}$};
    \path (1) edge (2);
    \path (2) edge (3);
    \path (1) edge [bend left] (3);
    \path (4) edge (5);
    \path (5) edge (6);
    \path (4) edge [bend left] (6);
    \path[blue] (7) edge (8);
    \path[blue] (8) edge (9);
    \path[blue] (7) edge [bend left] (9);
    \path (10) edge (11);
    \path (11) edge (12);
    \path (10) edge [bend left] (12);
    \path (13) edge (14);
    \path (14) edge (15);
    \path (13) edge [bend left] (15);
    \path (2) [dashed] edge (6);
    \path (5) [dashed] edge (3);
    \path (5) [dashed] edge (9);
    \path[blue] (8) [dashed] edge (6);
    \path[blue] (8) [dashed] edge (12);
    \path (11) [dashed] edge (9);
    \path (11) [dashed] edge (15);
    \path (14) [dashed] edge (12);
    \draw[blue,thick,opacity=0.3] ($(6.north west)+(-0.4,0.3)$)  rectangle ($(10.south east)+(0.4,-0.3)$);
    \node[blue] at (3,4.3){neighboring area};
    \end{tikzpicture}
    \caption{Local interference}
    \label{fig:1b}
\end{subfigure}
\caption{Causal relationships among the treatment $T$, the outcome $Y$, and the observed confounder $X$ (with $U$ excluded) under the assumptions of (a) no interference and (b) local interference where the neighboring area $\mathcal{S}$ only covers the immediate neighbors of each unit. Here we show a simplified 1-dimensional example with neighborhood window of size 1, although the spatial problem is 2-dimensional and neighborhood window is larger in the work presented here.}
\label{fig:1}
\vspace{-3mm}
\end{figure*}

We first define the necessary notations and assumptions for causal inference on spatial data. We will use upper-case letters (e.g., $T_1,\dots,T_M$, $Y$) for one-dimensional random variables and lower-case letters (e.g., $t_1,\dots,t_M$, $y$) for their corresponding realizations. Multi-dimensional random variables and realizations will be denoted using bold fonts (e.g., $\boldsymbol{X}$ and $\boldsymbol{x}$). In conventional causal inference settings with one single treatment, it is standard to define $T$ as the treatment, $\boldsymbol{X}$ as the set of observed confounders, and $Y$ as the outcome. However, in our case, we may have multiple treatments. If so, we denote them as $T_1, \dots, T_M$; otherwise, we use $T$ for a single treatment. As discussed in the introduction, in spatial settings, the treatment received by one unit can affect the outcome of other units, a phenomenon known as \emph{spatial interference} or \emph{spillover effect}. While in an ideal scenario, the outcome of one unit could be influenced by all units within a spatial region, we restrict our consideration to the set of treatments $\overline{\boldsymbol{T}}$ within a defined ``neighboring area'' (see Figure \ref{fig:1b}, which simplifies the problem to a one-dimensional case, though the actual problem is in two dimensions) for each unit, which we refer to as \emph{local interference}. We posit that treatments beyond this area are distant enough from the unit to exert a negligible impact on the outcome. In addition to these observed variables, as discussed in the introduction, we allow for an unobserved confounder $U$, which is assumed to be a purely spatial term and thus the same for all observations within a spatial region.

Point-referenced data are often measured at a unique geographical location rather than over a region. Let $\boldsymbol{s}_i \in \mathbb{R}^{d_s}$ be the spatial coordinates associated with each individual unit $i$, where $d_s$ denotes the dimensionality of the spatial coordinates. Following \cite{reich2021review}, under the assumption of no interference (see Figure \ref{fig:1a}), we can define the linear spatial regression model as
\begin{equation}
Y_i = \sum_{m=1}^{M} \alpha_m T_{i,m} + \boldsymbol{\gamma}^T \boldsymbol{X}_i + U(\boldsymbol{s}_i) + \varepsilon_i, 
\label{eq:1}
\end{equation}
where there are $M$ different treatments, $\alpha_m$ is the coefficient for the \emph{direct effect} of the $m^{th}$ treatment assigned to the unit, $\boldsymbol{\gamma}$ determines the effect of observed confounders, $U$ denotes the unobserved spatial random effect, and $\varepsilon \sim \mathcal{N}(0, \sigma_{\varepsilon}^2)$ is an i.i.d. random noise term. As mentioned previously, $U(\boldsymbol{s})$ is often assumed to be a continuous function over the entire spatial region and modeled as a GP with zero mean and isotropic covariance function. However, if we consider the local interference as shown in Figure \ref{fig:1b}, we need to introduce an extra term $\overline{T}$ to the regression model in Equation \ref{eq:1} to account for the \emph{indirect effect} from neighboring treatments, i.e.,
\begin{equation}
Y_i = \sum_{m=1}^{M} \alpha_m T_{i,m} + \sum_{m=1}^{M} f_{\overline{T}} \left( \overline{\boldsymbol{T}}_{i,m} \right) + \boldsymbol{\gamma}^T \boldsymbol{X}_i + U(\boldsymbol{s}_i) + \varepsilon_i,
\label{eq:2}
\end{equation}
where $\overline{\boldsymbol{T}}_{i,m}$ represents all treatments assigned to the units within the neighboring area $\mathcal{S}$ except for the current unit, i.e., $\overline{\boldsymbol{T}}_{i,m} = \{T_{i,m} (\bar{\boldsymbol{s}}_i) | \bar{\boldsymbol{s}}_i \in \mathcal{S}, \bar{\boldsymbol{s}}_i \neq \boldsymbol{s}_i\}$ and $f_{\overline{T}}$ is some function that models the \emph{indirect effect} from $\overline{\boldsymbol{T}}_{i,m}$. For example, \citet{reich2021review} state that $f_{\overline{T}}$ can be a weighted integral of the treatments within $\mathcal{S}$, excluding the spatial coordinates of the current unit.

\subsection{Potential Outcomes Framework with Continuous Treatments}
\label{sec:2.1}
In accordance with the potential outcomes framework \cite{rosenbaum1983central,rubin2005causal}, we first make the following assumptions to make sure that causal effects can be appropriately identified. Both scenarios involving binary and continuous treatments will be examined.

\begin{assumption}
\label{assp:1}
(Consistency) The observed outcome is the potential outcome determined by all the observed treatment assignments within $\mathcal{S}$, i.e., $Y = Y(T_1, ..., T_M, \overline{\boldsymbol{T}}_1, ..., \overline{\boldsymbol{T}}_M)$.
\end{assumption}

\begin{assumption}
\label{assp:2}
(Latent ignorability \cite{frangakis1999addressing}) The potential outcomes and treatments are independent given both the observed and the unobserved confounders, i.e., \textcolor{black}{$Y \independent T_m | \boldsymbol{X}, U \; \forall \; m \in \{1, ..., M\}$.}
\end{assumption}

\begin{assumption}
\label{assp:3}
(Positivity) The conditional probability of the treatment assignment (or the propensity score) is always positive, i.e., \textcolor{black}{$p(T_m = t_m|\boldsymbol{X},U) > 0 \; \forall \; m \in \{1, ..., M\}$ where $t_m \in \mathbb{R}$ for continuous treatments.}
\end{assumption}

Our goal is to estimate the direct effect (DE), the indirect effect (IE), and the total effect (TE) under local interference for all $M$ types of treatments as specified below. To simplify notations, we write $Y(T_m = t_m, \overline{\boldsymbol{T}}_m = \bar{\boldsymbol{t}}_m, T_{-m} = t_{-m}, \overline{\boldsymbol{T}}_{-m} = \bar{\boldsymbol{t}}_{-m}, \boldsymbol{X} = \boldsymbol{x}) = Y(t_m, \bar{\boldsymbol{t}}_m)$ where $T_{-m}$ and $\overline{\boldsymbol{T}}_{-m}$ denote all other types of treatment except the $m^{th}$ one assigned on the current unit and neighboring units within $\mathcal{S}$, respectively. Both $T_{-m}$ and $\overline{\boldsymbol{T}}_{-m}$ will be held at their observed values when calculating these effects. 

When all treatments are \emph{continuous}, i.e., $T_m \in \mathbb{R} \: \forall \: m \in \{1, ..., M\}$, it is common practice to calculate direct, indirect, and total effects as follows:
\begin{align}
\text{DE}(t_m, \bar{\boldsymbol{t}}_m) &= \mathbb{E}_{\mathcal{D}} \left[ Y(t_m, \bar{\boldsymbol{t}}_m) - Y(0, \bar{\boldsymbol{t}}_m) \right], \label{eq:3} \\
\text{IE}(t_m, \bar{\boldsymbol{t}}_m) &= \mathbb{E}_{\mathcal{D}} \left[ Y(t_m, \bar{\boldsymbol{t}}_m) - Y(t_m, \boldsymbol{0}) \right], \label{eq:4} \\
\text{TE}(t_m, \bar{\boldsymbol{t}}_m) &= \mathbb{E}_{\mathcal{D}} \left[ Y(t_m, \bar{\boldsymbol{t}}_m) - Y(0, \boldsymbol{0}) \right], \label{eq:5}
\end{align}
where, in practice, the expectation over the dataset $\mathbb{E}_{\mathcal{D}}$ can be calculated as the average difference between factual and counterfactual outcomes over all samples in $\mathcal{D}$. According to \citet{reich2021review}, there can be infinitely many different possible values for DE for continuous treatment, but here we only focus on the case where $\bar{\boldsymbol{t}}_m$ is set to observed values instead of calculating DE for all possible combinations of neighboring treatments within $\mathcal{S}$.

Furthermore, in (semi-)synthetic cases we can integrate over the treatment distribution as well.
Consequently, we draw upon the notion of the average dose-response function \cite{kreif2015evaluation} to formulate \emph{alternative} definitions for direct, indirect, and total effects as shown below:
\begin{align}
\text{DE}(\bar{\boldsymbol{t}}_m) &= \mathbb{E}_{\mathcal{D}} \left[ \mathbb{E}_{t_m} \left[ Y(t_m, \bar{\boldsymbol{t}}_m) - Y(0, \bar{\boldsymbol{t}}_m) \right] \right], \label{eq:6} \\
\text{IE}(t_m) &= \mathbb{E}_{\mathcal{D}} \left[ \mathbb{E}_{\bar{\boldsymbol{t}}_m} \left[ Y(t_m, \bar{\boldsymbol{t}}_m) - Y(t_m, \boldsymbol{0}) \right] \right], \label{eq:7} \\
\text{TE} &= \mathbb{E}_{\mathcal{D}} \left[ \mathbb{E}_{t_m, \bar{\boldsymbol{t}}_m} \left[ Y(t_m, \bar{\boldsymbol{t}}_m) - Y(0, \boldsymbol{0}) \right] \right], \label{eq:8}
\end{align}
where both $t_m$ and $\bar{\boldsymbol{t}}_m$ are drawn from the space of the $m^{th}$ treatment $\mathcal{T}_m$. In our experiments, we use \eqref{eq:6}-\eqref{eq:8} when the ground-truth dose-response function is known. 
Otherwise, we use \eqref{eq:3}-\eqref{eq:5}.

\section{Deep Spatial Causal Inference with Continuous Treatments}
\label{sec:3}

The core idea of our methodology is to model the interference and measured confounding effects using deep neural networks (NNs). This approach is motivated by the fact that in practical environmental applications, interference and confounding effects often exhibit nonlinear or non-stationary patterns \cite{li2020spatially,rollinson2021working}. Consequently, traditional linear or stationary models may fail to adequately capture these complexities. To address this, we modify the spatial regression model presented in \eqref{eq:2} as follows:
\begin{equation}
\begin{split}
Y_i = &\sum_{m=1}^{M} \alpha_m T_{i,m} + \sum_{m=1}^{M} f_{\theta_{t_m}} (\overline{\boldsymbol{T}}_{i,m}) + g_{\theta_x} (\boldsymbol{X}_i) + \\ &U_{\phi}(\boldsymbol{s}_i) + \varepsilon,
\label{eq:9}
\end{split}
\end{equation}
where both $f_{\theta_{t_m}}$ and $g_{\theta_x}$ are deep neural networks parameterized by $\theta_{t_m}$ and $\theta_x$, respectively, and $U_{\phi}(\boldsymbol{s}_i)$ is a sample drawn from an approximate GP parameterized by $\phi$. Note that here we keep $T_{i,m}$ as a linear term to ensure that the direct effect of $T_{i,m}$ is not overshadowed by the indirect effect of $\overline{\boldsymbol{T}}_{i,m}$, particularly when $\overline{\boldsymbol{T}}_{i,m}$ has a much higher dimensionality than $T_{i,m}$. If desired, a neural network could be used on this term as well. In addition, to appropriately handle partially observed outcomes, we adopt a generalized propensity-score-based approach to calculate \emph{balancing weights} when estimating causal effects. We will explain the network architectures, the unobserved confounder $U_{\phi}$, and the generalized propensity-score-based in more detail in the following sections.

\subsection{Network Architectures}
\label{sec:3.1}

Ideally, a wide range of network architectures can be considered for $f_{\theta_{t_m}}$ and $g_{\theta_x}$. In this context, we select from the following types of architectures for $f_{\theta_{t_m}}$ and $g_{\theta_x}$ depending on the type of inputs:

\begin{itemize} [leftmargin=*]

\item \textbf{Multi-layer perceptron (MLP)} for $g_{\theta_x}$: Fully connected neural networks with multiple hidden layers and nonlinear activation functions are natural choices for modeling nonlinear functional relationships. In the context of spatial causal inference, where the measured confounder $\boldsymbol{X}$ is typically represented as a vectorized input, we can set $g_{\theta_x}$ as an MLP to capture potential nonlinear relationships between $\boldsymbol{X}$ and $Y$.

\item \textbf{Convolutional neural network (CNN)} for $f_{\theta_{t_m}}$: In the previous section, we defined a neighboring area $\mathcal{S}$, where all units except the central one contribute to local interference or indirect effect. This definition is somewhat similar to the concept of a \emph{receptive field} in a CNN. Additionally, the neighboring treatments $\overline{\boldsymbol{T}}_m$ are usually represented by a two-dimensional matrix in spatial data. Therefore, it is intuitive to employ a CNN to model the indirect effects of  $\overline{\boldsymbol{T}}_m$ on $Y$.

\item \textbf{U-Net} for $f_{\theta_{t_m}}$: U-Net \cite{ronneberger2015u} is a convolutional neural network (CNN) architecture characterized by its dual-path structure: a contracting path and an expanding path. 
As emphasized by \citet{tec2023weather2vec}, a noteworthy characteristic of U-Net is its ability to transform an input spatial map into an output spatial map, where each element in the output localizes contextual spatial information from the input.

\end{itemize}

\subsection{Modeling the Unobserved Confounder as a sample from an Approximate GP}
\label{sec:3.2}

We leverage the Implicit Composite Kernel (ICK) framework proposed by \citet{jiang2024incorporating} to model the unobserved confounder, which can be viewed as a sample drawn from an approximate GP with weighted inducing points. To elaborate, we first take the spatial coordinates $\boldsymbol{s}$ and compute the covariance matrix $\boldsymbol{K}$ where the entries of $\boldsymbol{K}$ can be calculated as $\boldsymbol{K}_{ij} = K_{\varphi}(\boldsymbol{s}_i, \boldsymbol{s}_j)$ and $K_{\varphi}(\cdot, \cdot)$ is an isotropic covariance function with parameter $\varphi$. Next, we approximate the covariance matrix $\boldsymbol{K}$ using a low-rank matrix with $q$ inducing points, i.e., $\boldsymbol{K}_q \in \mathbb{R}^{q \times q}$, yielding $\boldsymbol{K} \approx \hat{\boldsymbol{K}} = \boldsymbol{K}_{nq} \boldsymbol{K}_q^{-1} \boldsymbol{K}_{nq}^{T}$. The entries of $\boldsymbol{K}_q$ and $\boldsymbol{K}_{nq}$ can be calculated as $\left(\boldsymbol{K}_q\right)_{ij} = K_{\varphi}(\hat{\boldsymbol{s}}_i, \hat{\boldsymbol{s}}_j), i, j \in \{1, 2, ..., q\}$ and $\left(\boldsymbol{K}_{nq}\right)_{ij} = K_{\varphi}(\boldsymbol{s}_i, \hat{\boldsymbol{s}}_j), i\in\{1, 2, ..., N\}, j\in\{1, 2, ..., q\}$, respectively. We then perform Cholesky decomposition $\boldsymbol{K}_q^{-1} = \boldsymbol{U}^T \boldsymbol{U}$ and calculate the vector $\boldsymbol{z}_i \triangleq \boldsymbol{U}(\boldsymbol{K}_{nq}^T)_{:, i} \: \forall \: i = 1, 2, ..., N$. Finally, the effect of the unobserved confounder on the target $Y$ can be modeled as $U_{\phi}(\boldsymbol{s}_i) = \boldsymbol{w}^T \boldsymbol{z}_i$ where $\boldsymbol{w}$ is a vector of learnable weights. Thus, we have $\phi = \{\boldsymbol{w}, \varphi\}$. Based on both theoretical statements and empirical evidence presented in \citet{jiang2024incorporating}, we can efficiently approximate the behavior of a GP with covariance function $K_{\varphi}$ using this method.

With the approach described above, we successfully achieve the \emph{joint} learning of NNs and an approximate GP using stochastic gradient descent, as both low-rank matrix approximation and Cholesky decomposition are differentiable. To the best of our knowledge, no prior work has combined NNs and GPs in this manner.  As previously mentioned, when the scale of unobserved spatial confounding is greater than the spatial variation of the observed variables, including this effect reduces the bias of estimates of treatment effects \cite{paciorek2010importance}. This innovative approach not only allows for the integration of NNs and GPs within a unified framework but also enhances the model's capacity to capture complex spatial dependencies in the unobserved confounder, as we will demonstrate empirically.

\subsection{Addressing Partially Observed Outcomes}
\label{sec:3.3}

An important consideration in the definitions of direct, indirect, and total effects in \eqref{eq:6}-\eqref{eq:8} is the need to take the expectation over the entire treatment space $\mathcal{T}$ for each sample in our dataset. However, in real-world or semi-synthetic settings, we typically observe the outcome for only one specific treatment assignment per sample, i.e., $T_m = t_m$. Consequently, the outcomes for each sample in our dataset are partially observed. This scenario is somewhat analogous to the bandit problem in reinforcement learning, except that we do not assign treatments iteratively and, theoretically, there are infinite possible treatment values.

To address the issue of partially observed outcomes, we follow a similar approach in \citet{robins2000marginal} and \citet{assaad2021counterfactual} and apply balancing weights $\boldsymbol{w}$ to all the predicted outcomes $\hat{y}$ when calculating the direct, indirect, and total effects. Specifically, for the $i^{th}$ sample in the dataset, to evaluate the direct, indirect, and total effects for the $m^{th}$ treatment, the balancing weight is
\begin{equation}
\textstyle
\boldsymbol{w}_{m}^{(i)} \coloneqq \frac{f_{T_m}\left( 
T_m^{(i)} = t_m^{(i)} \right)}{f_{T_m | \boldsymbol{X}, \boldsymbol{s}}\left( T_m^{(i)} = t_m^{(i)} | \boldsymbol{X}^{(i)}, \boldsymbol{s}^{(i)} \right)}, \label{eq:10}
\end{equation}
where $f_{T_m}\left( 
T_m^{(i)} = t_m^{(i)} \right)$ is the marginal density and $f_{T_m | \boldsymbol{X}, \boldsymbol{s}}\left( T_m^{(i)} = t_m^{(i)} | \boldsymbol{X}^{(i)}, \boldsymbol{s}^{(i)} \right)$ is the generalized propensity score (GPS) \cite{hirano2004propensity} of the $m^{th}$ treatment for the $i^{th}$ sample. It is important to note that the GPS extends the concept of the traditional propensity score, which is primarily designed for binary treatments, to accommodate continuous or multi-level treatments. In our experiments, the marginal density is estimated using kernel smoothing, and the GPS is estimated using a logistic regression model. Note that GPS is not conditioned on neighboring treatments $\overline{\boldsymbol{T}}$ as we do not assume any spatial relationships among treatment variables as depicted in Figure \ref{fig:1}. In addition, to avoid propensity overfitting \cite{swaminathan2015self}, we further weigh each predicted outcome by a self-normalization term $\frac{1}{N} \sum_{i=1}^{N} \boldsymbol{w}_{m}^{(i)}$ where $N$ is the total number of samples in our dataset for causal effect estimation.

\section{Related Work}
\label{sec:4}

To date, various approaches have been developed to address spatial interference and missing confounder in causal inference within spatial settings \cite{akbari2023spatial}. The concept of a ``neighboring area'', or ``clique'', as discussed Section \ref{sec:2}, was initially introduced by \citet{besag1974spatial}, with subsequent works adopting similar frameworks. For instance, \citet{delgado2015difference} employed a difference-in-difference regression method to explicitly account for indirect effects from proximate neighbors. \citet{aronow2017estimating} estimated causal effects under spatial interference by defining an exposure mapping that associates treatment assignment vectors with exposure values received by units in localized regions. Besides these, \citet{forastiere2021identification} developed covariate-adjustment methods to produce valid estimates of direct and indirect effects, relying on an extended unconfoundedness assumption. Moreover, \citet{giffin2023generalized} proposed a Bayesian spline-based regression model to recover direct and indirect effects using a generalized propensity score approach. On the other hand, common methods for dealing with unobserved spatial confounder include matching methods \cite{giudice2019selection,papadogeorgou2019adjusting,kolak2020spatial,giffin2023generalized}, difference-in-difference \cite{delgado2015difference,bardaka2019spatial}, structural equation models \cite{wright1934method,thaden2018structural}, instrumental variables \cite{marcos2019economic,giffin2021instrumental}, and regression discontinuity design \cite{imbens2008regression,bor2014regression,keele2015geographic,d2015housing}.

\begin{wraptable}{R}{0.65\textwidth}
\vspace{-4mm}
\setlength{\tabcolsep}{0.3em}
\caption{Estimation errors of the direct, indirect, and total effects on the synthetic toy dataset for both linear and NN-based spatial regression models.}
\label{tab:1}
\centering
\resizebox{0.65\textwidth}{!}{
\begin{tabular}{l|cccc} 
\hline
\begin{tabular}[c]{@{}l@{}}\end{tabular}    & \begin{tabular}[c]{@{}c@{}}Linear model\\~(without $U_{\phi}$)\end{tabular} & \begin{tabular}[c]{@{}c@{}}Linear model\\~(with $U_{\phi}$)\end{tabular} & \begin{tabular}[c]{@{}c@{}}NN-based model\\~(without $U_{\phi}$)\end{tabular} & \begin{tabular}[c]{@{}c@{}}NN-based model\\~(with $U_{\phi}$)\end{tabular}  \\ 
\hline
\begin{tabular}[c]{@{}l@{}}Direct \\effect (DE)\end{tabular}   & \begin{tabular}[c]{@{}c@{}}0.709\\~$\pm$ 0.023\end{tabular}          & \begin{tabular}[c]{@{}c@{}}0.108 \\$\pm$ 0.012\end{tabular}       & \begin{tabular}[c]{@{}c@{}}\textbf{0.078} \\$\pm$ \textbf{0.049}\end{tabular}            & \begin{tabular}[c]{@{}c@{}}0.096 \\$\pm$ 0.016\end{tabular}          \\
\begin{tabular}[c]{@{}l@{}}Indirect~\\effect (IE)\end{tabular} & \begin{tabular}[c]{@{}c@{}}0.806\\~$\pm$ 0.080\end{tabular}          & \begin{tabular}[c]{@{}c@{}}0.123 \\$\pm$ 0.010\end{tabular}       & \begin{tabular}[c]{@{}c@{}}0.231 \\$\pm$ 0.054\end{tabular}            & \begin{tabular}[c]{@{}c@{}}\textbf{0.116} \\$\pm$ \textbf{0.005}\end{tabular}          \\
\begin{tabular}[c]{@{}l@{}}Total \\effect (TE)\end{tabular}    & \begin{tabular}[c]{@{}c@{}}0.781\\~$\pm$ 0.071\end{tabular}          & \begin{tabular}[c]{@{}c@{}}0.115 \\$\pm$ 0.018\end{tabular}       & \begin{tabular}[c]{@{}c@{}}0.347 \\$\pm$ 0.093\end{tabular}            & \begin{tabular}[c]{@{}c@{}}\textbf{0.095} \\$\pm$ \textbf{0.016}\end{tabular}          \\
\hline
\end{tabular}}
\end{wraptable}

However, most of the existing spatial causal inference methodologies only work for simple vectorized features and thus do not adapt to high-dimensional inputs. While efforts have been made over the past decade to address this limitation through neural networks, such endeavors remain relatively sparse. For example, \citet{pollmann2020causal} identified counterfactual treatment locations on a spatial map with discretized latitude and longitude using CNNs, which facilitates the causal effect estimation in scenarios where treatments exhibit spatial patterns. \citet{tec2023weather2vec} developed a U-net-based \cite{ronneberger2015u} framework to learn representations of non-local information for each observational unit. This approach is particularly useful in situations where treatments and outcomes for a given unit are influenced by measured confounders of nearby units, known as non-local confounding. Compared to these studies, our method addresses spatial interference and unobserved confounding by incorporating spatial information into neighboring treatments and missing spatial confounders. While we do not explicitly assume the existence of non-local confounding in the scope of this work, we believe it can be partially accommodated if we further include confounder features of neighboring units in $\boldsymbol{X}$. In addition, although a variety of deep learning-based frameworks exist for non-spatial causal inference \cite{shalit2017estimating,shi2019adapting,hatt2021estimating,jiang2023estimating,jiang2023causal}, adapting them to spatial settings remains an open challenge.

\section{Simulation on Synthetic Data}
\label{sec:5}

We first test our proposed methods using a synthetic toy dataset constructed on a one-dimensional line graph (i.e., $\boldsymbol{s} \in \mathbb{R}$) comprising 500 nodes where each node is characterized by variables including treatment $T$, observed confounder $\boldsymbol{X}$, unobserved confounder $U$, and outcome $Y$. Note that we only have one treatment so $m = 1$ in this case. The model for the outcome of the $i^{th}$ node can be expressed as:
\begin{equation}
Y_i = \beta T_i + f_T \left( \boldsymbol{d}_i \odot \overline{\boldsymbol{T}}_i \right) + f_X (\boldsymbol{X}_i) + U_i + \varepsilon, 
\label{eq:11}
\end{equation}
and $\boldsymbol{X}$, $U$, and $T$ are modeled as $\boldsymbol{X}_i \sim \mathcal{N}\left( \boldsymbol{0}, \sigma_X^2 \textbf{I} \right)$, $\boldsymbol{U} = \{U_i\}_{i=1}^{N} \sim \mathcal{N} \left( \boldsymbol{0}, \boldsymbol{D} \right)$ and $T_i = g \left( \boldsymbol{X}_i, U_i \right)$. Here we set $\textbf{I} \in \mathbb{R}^{4 \times 4}$ (i.e., $\boldsymbol{X}$ is 4-dimensional in this case) and $\boldsymbol{D} \in \mathbb{R}^{N \times N}$. Each element in $\boldsymbol{D}$ is calculated as $d_{ij} = \frac{1}{\sigma_d \sqrt{2 \pi}} \exp \left( -\frac{|s_i - s_j|}{2\sigma_l^2} \right)$ where $s_i, s_j \in [0,1]$ are the coordinates of the $i^{th}$ and $j^{th}$ nodes on the line graph, respectively, and $\boldsymbol{d}_i$ denotes the $i^{th}$ row of $\boldsymbol{D}$. We use a neighborhood size of 2 so that the outcome of each node is associated with its own treatment and its 2 adjacent nodes. Namely, $\overline{\boldsymbol{T}}_i$ contains only 2 nodes. $f_T$, $f_X$, and $g$ are all nonlinear functions and implemented as randomly initialized neural networks. The coefficient $\beta$ is drawn from a uniform distribution over $[0,1)$, while other parameters are fixed at $\sigma_X = 1$, $\sigma_d = 0.5$, and $\sigma_l = 0.5$. 

We fit our neural-network-based spatial regression model on the training data and compare the results with those obtained from a linear spatial regression model specified in \eqref{eq:2}. The function $f_{\overline{T}}$ in \eqref{eq:2} is implemented as a weighted sum of all treatment values in $\overline{\boldsymbol{T}}$, with all weights being trainable, in accordance with the methodology introduced by \citet{reich2021review}. Given the one-dimensional spatial structure of the data and the small neighborhood size, it is unnecessary to employ a convolutional architecture for $f_{\theta_t}$. Hence, both $f_{\theta_t}$ and $g_{\theta_x}$ in \eqref{eq:9} are implemented as MLPs with two hidden layers. All models are trained using a fixed learning rate of $0.001$, and the causal effects are estimated using the previously introduced balancing weight approach. For models without including the unobserved confounder, i.e., $U_{\phi}(\boldsymbol{s})$ in \eqref{eq:9}, we opt for stochastic gradient descent with a momentum of 0.99 as the optimizer. For models with the unobserved confounder, we use the Adam optimizer \citep{kingma2014adam} to ensure the positive-definiteness of the kernel matrix throughout the training process.

We repeat the experiment 5 times and present the mean and standard deviation of the estimation error on DE, TE, and TE based on \eqref{eq:6}-\eqref{eq:8} for each model in Table \ref{tab:1}. The results indicate that incorporating the approximate GP $U_{\phi}$ significantly improves the performance of both linear and NN-based models in estimating the direct, indirect, and total effects. Furthermore, the NN-based models consistently outperform the linear models. These findings align with our methodology, which uses nonlinear functions to simulate interference and confounding effects.

\section{Experiments on Geospatial Data}
\label{sec:6}

To evaluate our NN-based spatial regression framework in geospatial applications, we apply our model to Durham, North Carolina, using the heat island data collected on July 23, 2021, consistent with the study conducted by \citet{calhoun2024estimating}. Specifically, our analysis includes four datasets: the normalized difference vegetation index (NDVI) derived from Sentinel-2 imagery, an albedo dataset derived from Sentinel-2 observations, the National Land Cover Database (NLCD), and a surface air temperature dataset collected along a traversal path. These datasets are depicted in Figure \ref{fig:2}, with further details available in the appendix.

\begin{table*}[t!]
\centering
\begin{minipage}[c]{0.75\textwidth}
\setlength{\tabcolsep}{0.7em}
\begin{tabular}{cccc|ccc} 
\hline
$f_{\theta_{t_1}}$      & $g_{\theta_x}$      & $U_{\phi}$      & \begin{tabular}[c]{@{}c@{}}Balancing\\weight\end{tabular} & \begin{tabular}[c]{@{}c@{}}Direct\\effect (DE)\end{tabular} & \begin{tabular}[c]{@{}c@{}}Indirect\\effect (IE)\end{tabular} & \begin{tabular}[c]{@{}c@{}}Total\\effect (TE)\end{tabular}  \\ 
\hline
Linear & Linear &        &                  & 0.6919        & 1.6724          & 1.8844        \\
Linear & Linear &        & \checkmark           & 0.7453        & 1.7197          & 1.6385        \\
Linear & Linear & \checkmark &                  & 0.7685        & 1.7415          & 1.9612        \\
Linear & Linear & \checkmark & \checkmark           & \textbf{0.5635}        & 1.6936          & 1.6110        \\ 
\hline
CNN    & MLP    &        &                  & 0.8547        & 0.4351          & 0.0359        \\
CNN    & MLP    &        & \checkmark           & 0.6653        & 0.0765          & 0.0723        \\
CNN    & MLP    & \checkmark &                  & 0.7298        & 0.3454          & \textbf{0.0349}        \\
CNN    & MLP    & \checkmark & \checkmark           & 0.6280        & 0.0633          & 0.0694        \\
U-Net  & MLP    &        &                  & 0.7764        & 0.2827          & 0.2862        \\
U-Net  & MLP    &        & \checkmark           & 0.6706        & 0.2730          & 0.2997        \\
U-Net  & MLP    & \checkmark &                  & 0.8354        & 0.3469          & 0.2480        \\
U-Net  & MLP    & \checkmark & \checkmark           & 0.8564        & \textbf{0.0480}          & 0.1475        \\
\hline
\end{tabular}
\end{minipage}\hfill
\begin{minipage}[c]{0.25\textwidth}
\centering
\caption{Estimation errors of the direct, indirect, and total effects on the semi-synthetic geospatial data for both linear and NN-based spatial regression models with different types of NN architectures. Check marks on $U_{\phi}$ and balancing weight indicates that we include the GP term $U_{\phi}(\boldsymbol{s})$ and the balancing weight as defined in Equation \ref{eq:10} for estimating causal effects, respectively.}
\label{tab:2}
\end{minipage}
\end{table*}

\begin{figure}[t!]
\centering
\subfloat[\label{fig:2a}NDVI]{
\includegraphics[width=0.48\linewidth]{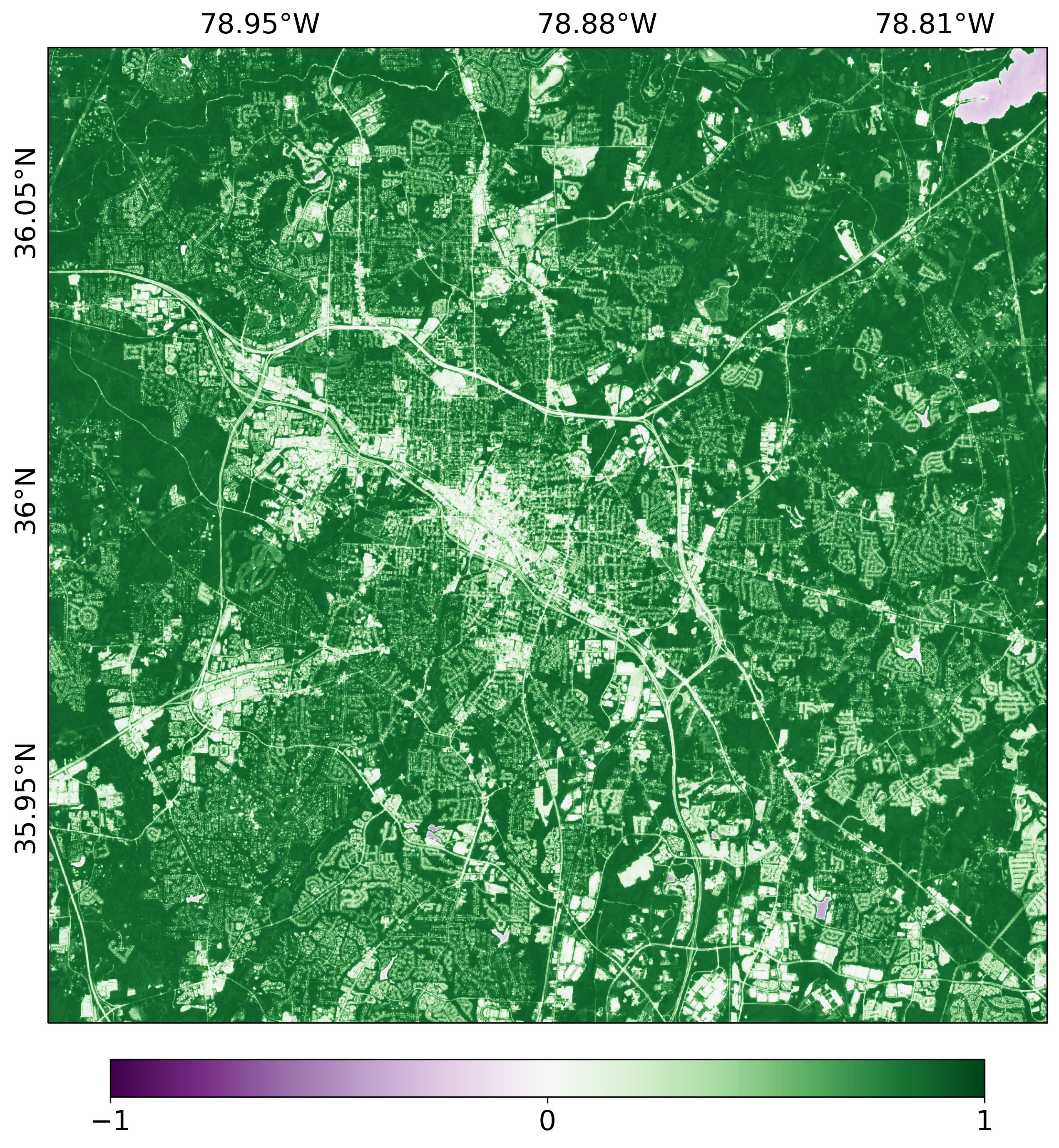}}
\hfill
\subfloat[\label{fig:2b}Albedo]{
\includegraphics[width=0.48\linewidth]{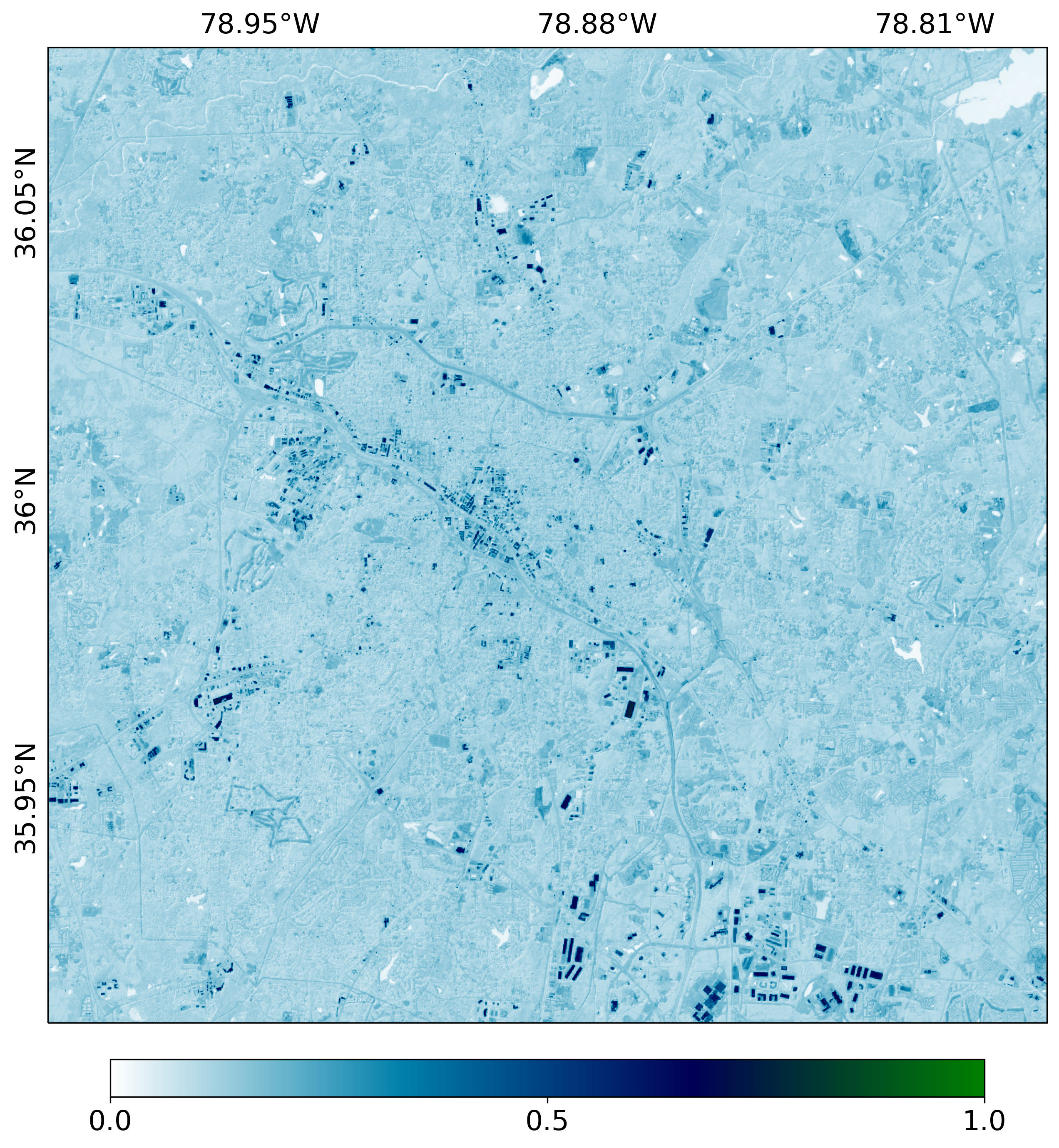}}
\vfill
\subfloat[\label{fig:2c}NLCD land cover classes]{
\includegraphics[width=0.48\linewidth]{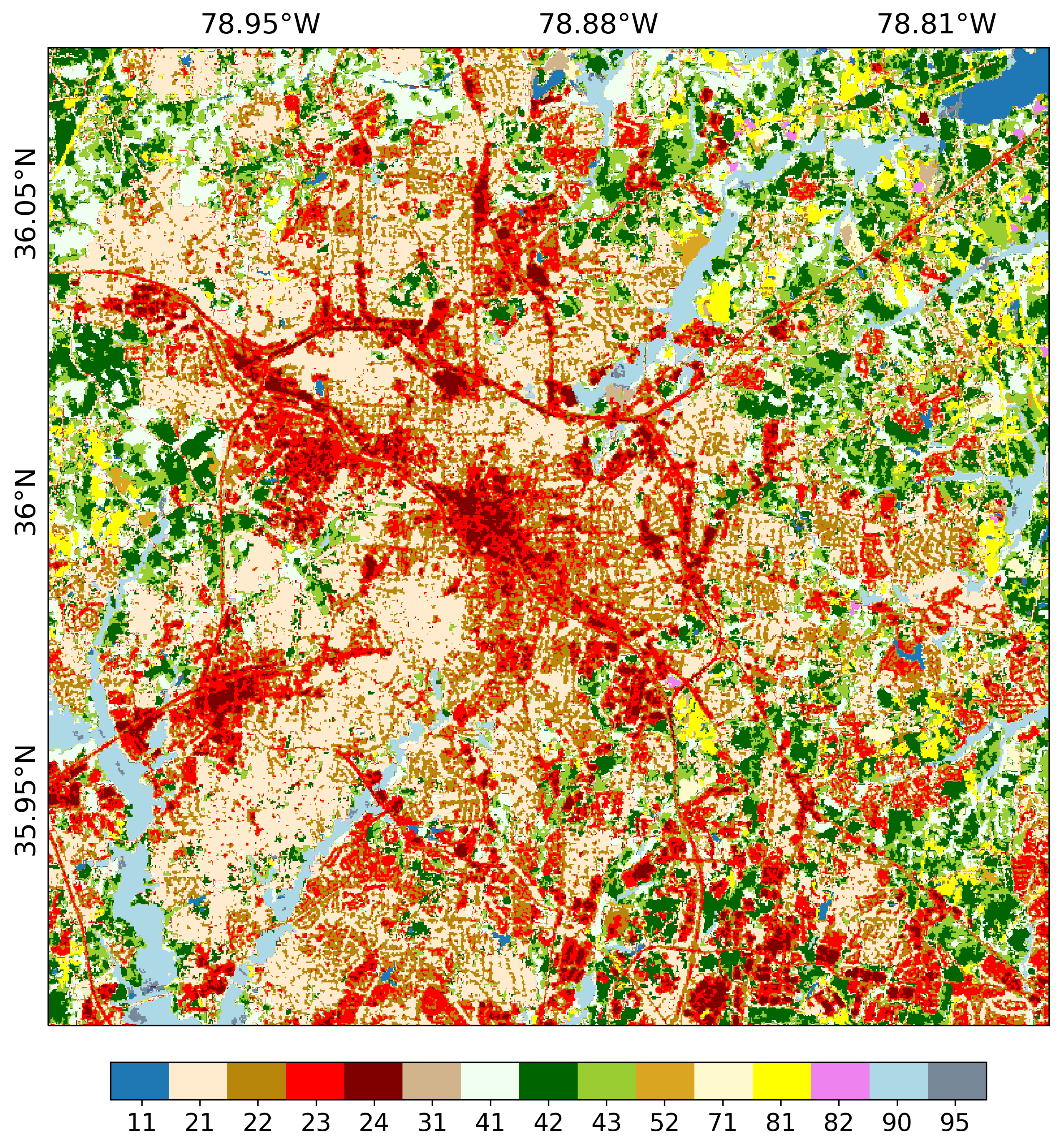}}
\hfill
\subfloat[\label{fig:2d}Traversal route and temperature (in Celsius)]{
\includegraphics[width=0.48\linewidth]{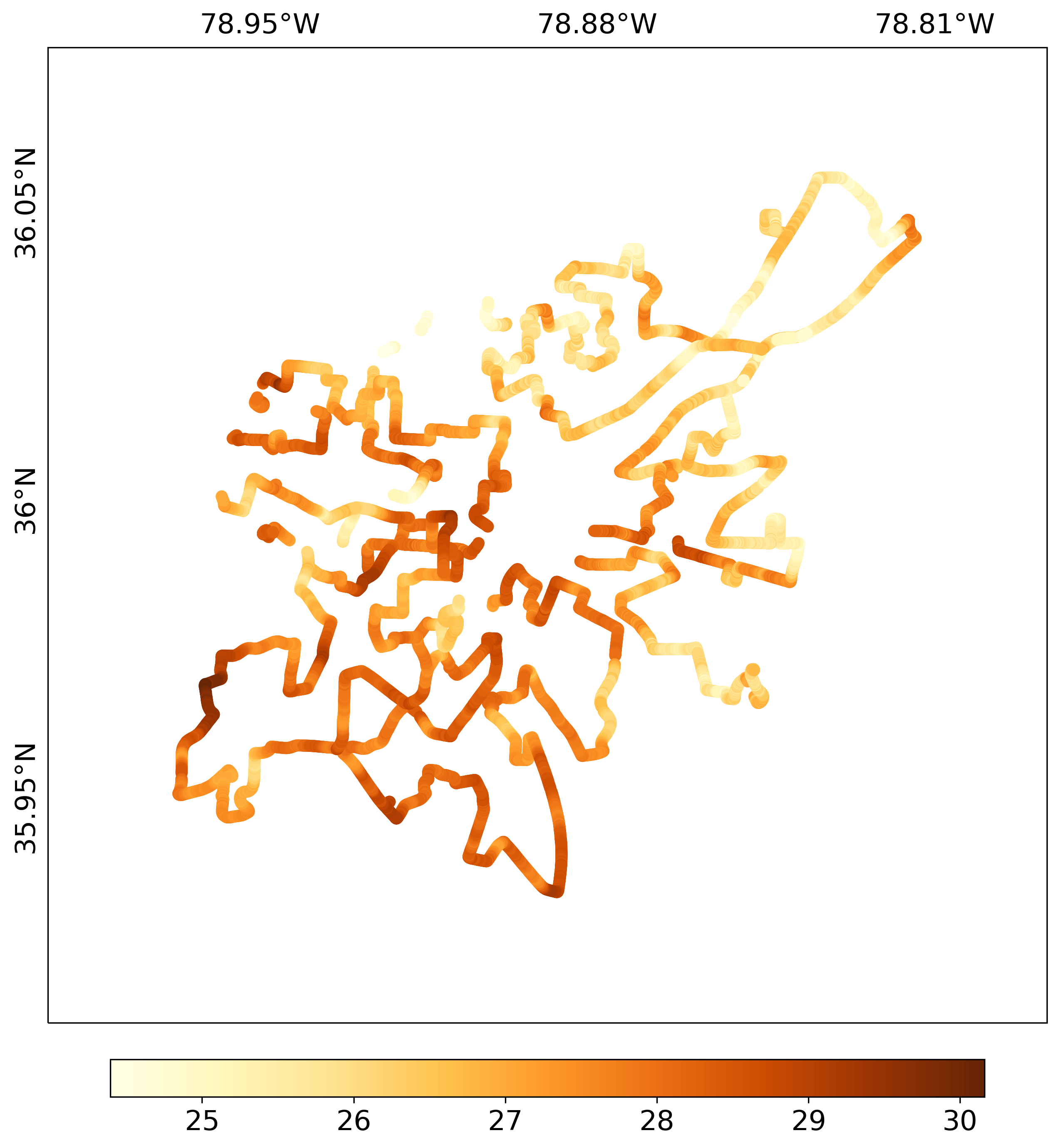}}
\caption{Geospatial datasets used for semi-synthetic and real-world experiments. Land cover classes are provided in Appendix \ref{appx:B}}
\label{fig:2}
\end{figure}

\subsection{Experiment on Semi-synthetic Data}
\label{sec:6.1}

We first test our proposed method on a semi-synthetic dataset where we include just the NDVI and NLCD for our analysis. In this context, NDVI is considered as the only treatment $T$, while NLCD serves as the observed confounder $\boldsymbol{X}$. Furthermore, we introduce an unobserved confounder $U$ \textcolor{blue}{sampled} from a GP with exponential kernel. To obtain \emph{synthetic} ground truth of both factual and counterfactual temperature values, we employ the following nonlinear model:
\begin{equation}
Y_i = \beta T_i + f_T(\boldsymbol{w}_i \odot \overline{\boldsymbol{T}}_i) + f_X (\boldsymbol{X}_i) + U_i, 
\label{eq:12}
\end{equation}
where $\boldsymbol{w}_i$ is a weight matrix over a neighboring area $\mathcal{S}$ with dimension $d_{\mathcal{S}} \times d_{\mathcal{S}}$, \textcolor{blue}{$\boldsymbol{s} \in \mathcal{S} \subset \mathbb{R}^2$}, and each element in the weight matrix is calculated as $w_{kl} = \frac{1}{Z} \exp \left( -d_{kl} / \sigma_l \right)$ with $Z$ being a normalization constant and $d_{kl}$ being the Euclidean distance between the pixel at position $(k,l)$, i.e., $0 \leq k, l < d_{\mathcal{S}}$, and the center pixel. Here the direct effect coefficient, the neighborhood size,  and the lengthscale parameter are set to be $\beta = -4$, $d_{\mathcal{S}} = 51$, and $\sigma_l = 10$, respectively. $f_T$ and $f_X$ are nonlinear functions generated by a piecewise cubic spline and a randomly initialized neural network, respectively.

Following the generation of semi-synthetic data, we divide it into training, validation, and test sets at a ratio of 6:2:2. It is important to note that the unobserved confounder $U$ is only used in data generation and is not included as part of the model input. Subsequently, we trained our NN-based spatial regression model on the training data, optimizing until convergence on the validation set using stochastic gradient descent with a fixed learning rate of $10^{-6}$. We then compared its performance on direct, indirect, and total effects again based on \eqref{eq:6}-\eqref{eq:8} with the linear counterparts of the model as defined in Equation \ref{eq:2}. The details of model architecture of $f_{\theta_{t_1}}$, $g_{\theta_x}$, and $U_{\phi}$ are provided in the appendix. As shown in Table \ref{tab:2}, the NN-based model significantly outperforms linear spatial regression models in estimating the indirect and total effects, while demonstrating comparable performance in estimating the direct effect. Additionally, we observe that incorporating balancing weights and the approximate Gaussian process $U_{\phi}$ further enhances the model's accuracy in estimating the indirect and total effects.

\subsection{Case Study on Real-world Data}
\label{sec:6.2}

\begin{figure*}[t!]
\centering
\subfloat[\label{fig:3a}Linear model]{
\includegraphics[width=0.245\linewidth]{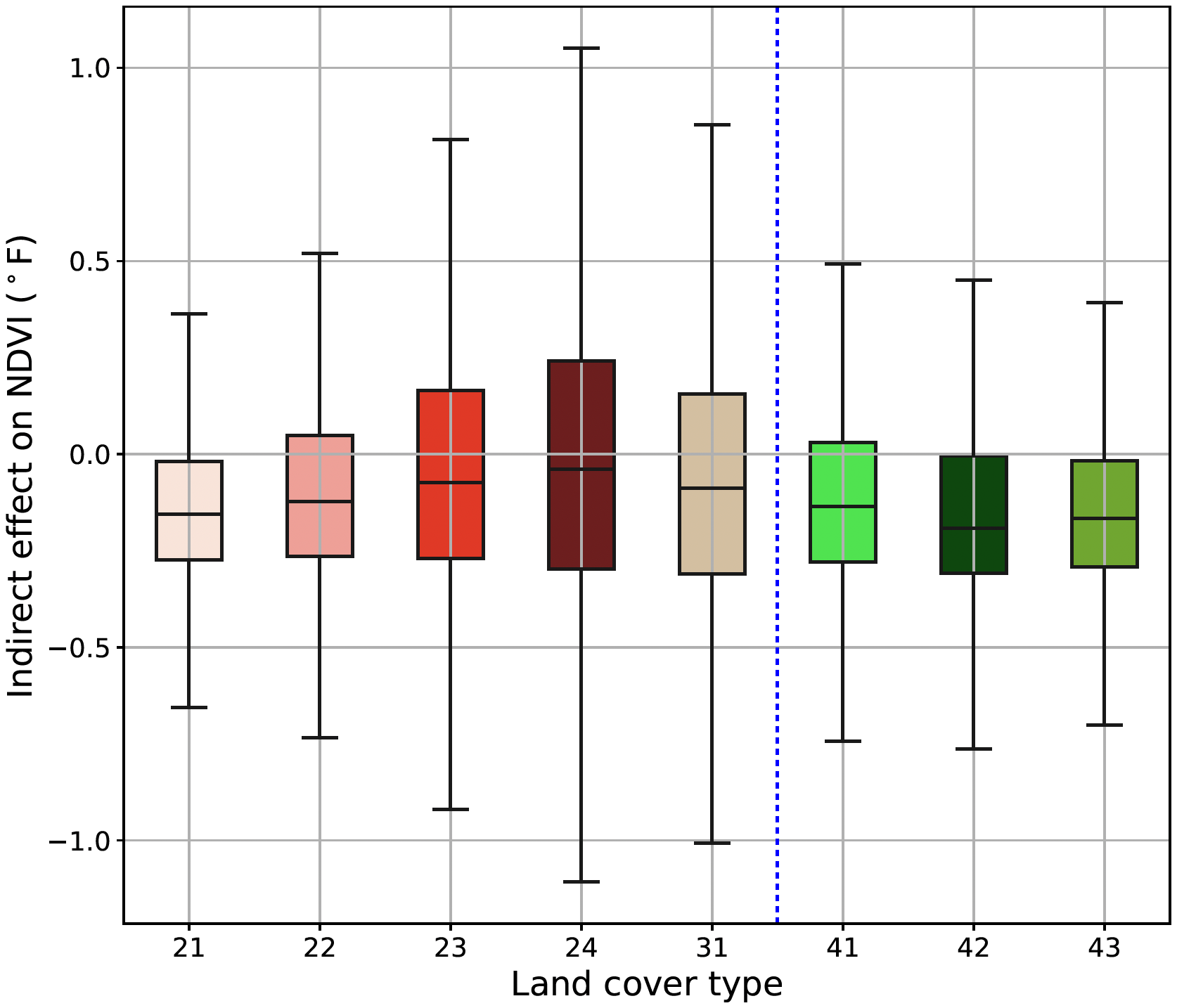}}
\hfill
\subfloat[\label{fig:3b}CNN-based model]{
\includegraphics[width=0.24\linewidth]{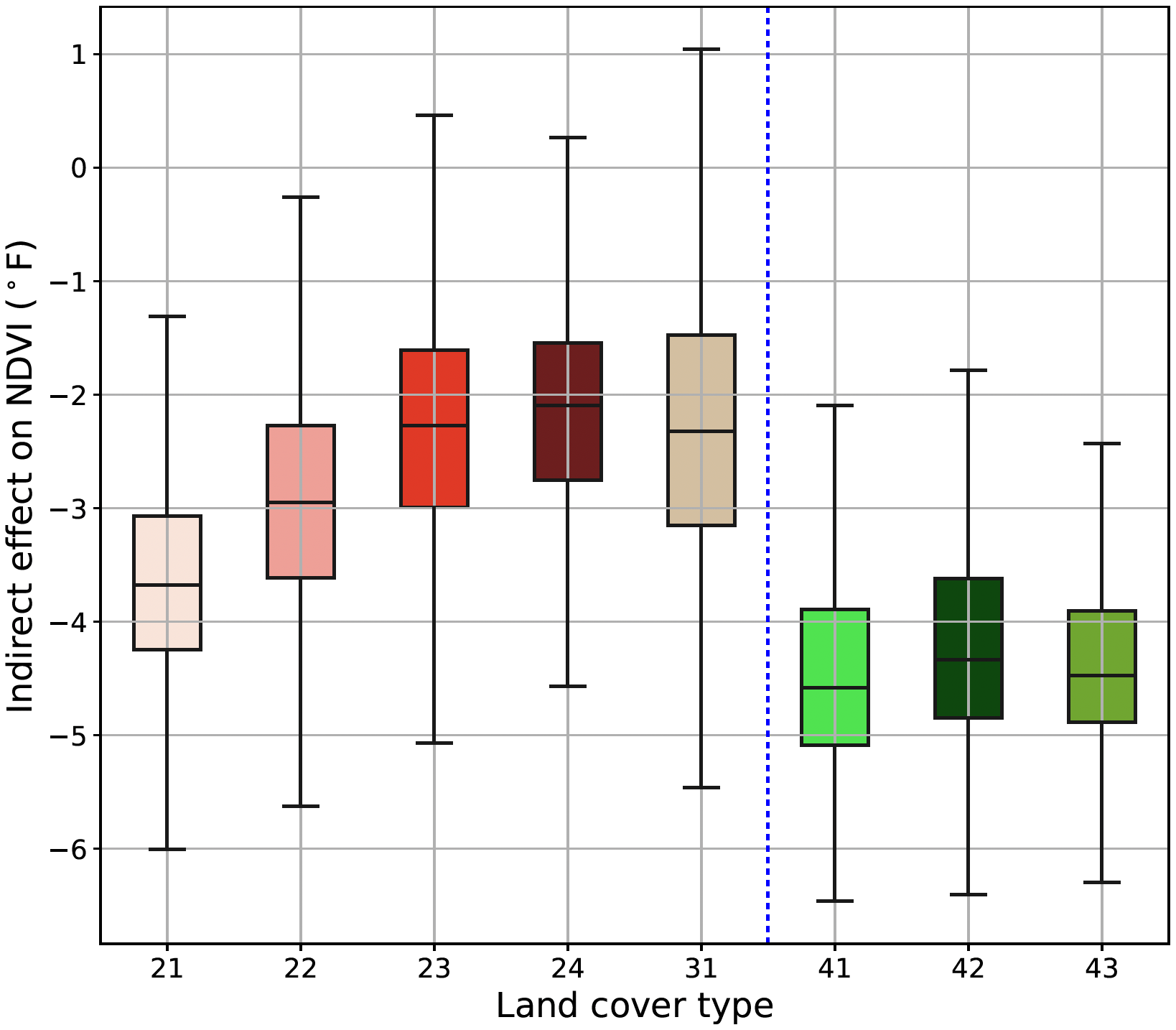}}
\hfill
\subfloat[\label{fig:3c}U-Net-based model]{
\includegraphics[width=0.245\linewidth]{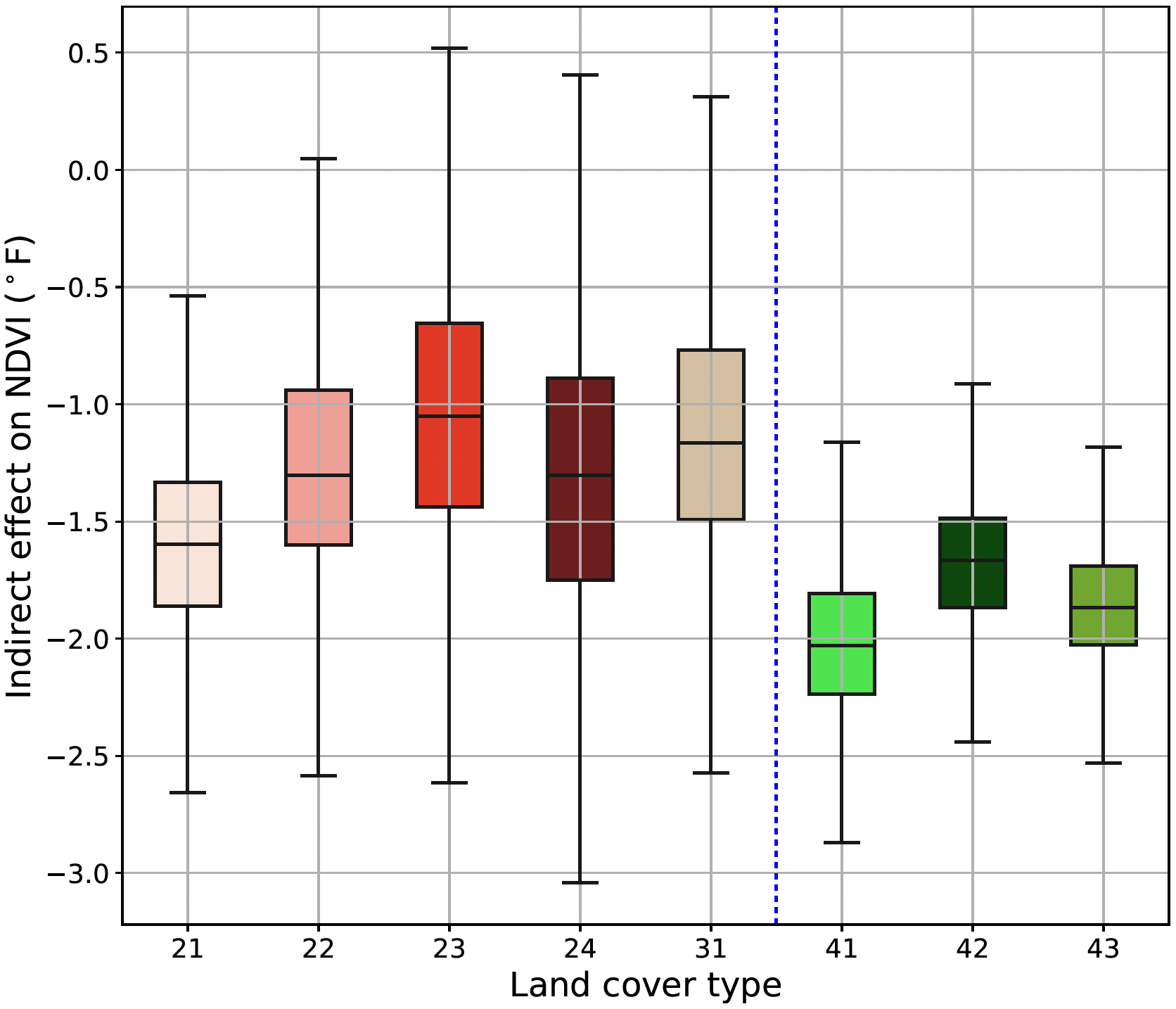}}
\hfill
\subfloat[\label{fig:3d}Ground truth measurements]{
\includegraphics[width=0.24\linewidth]{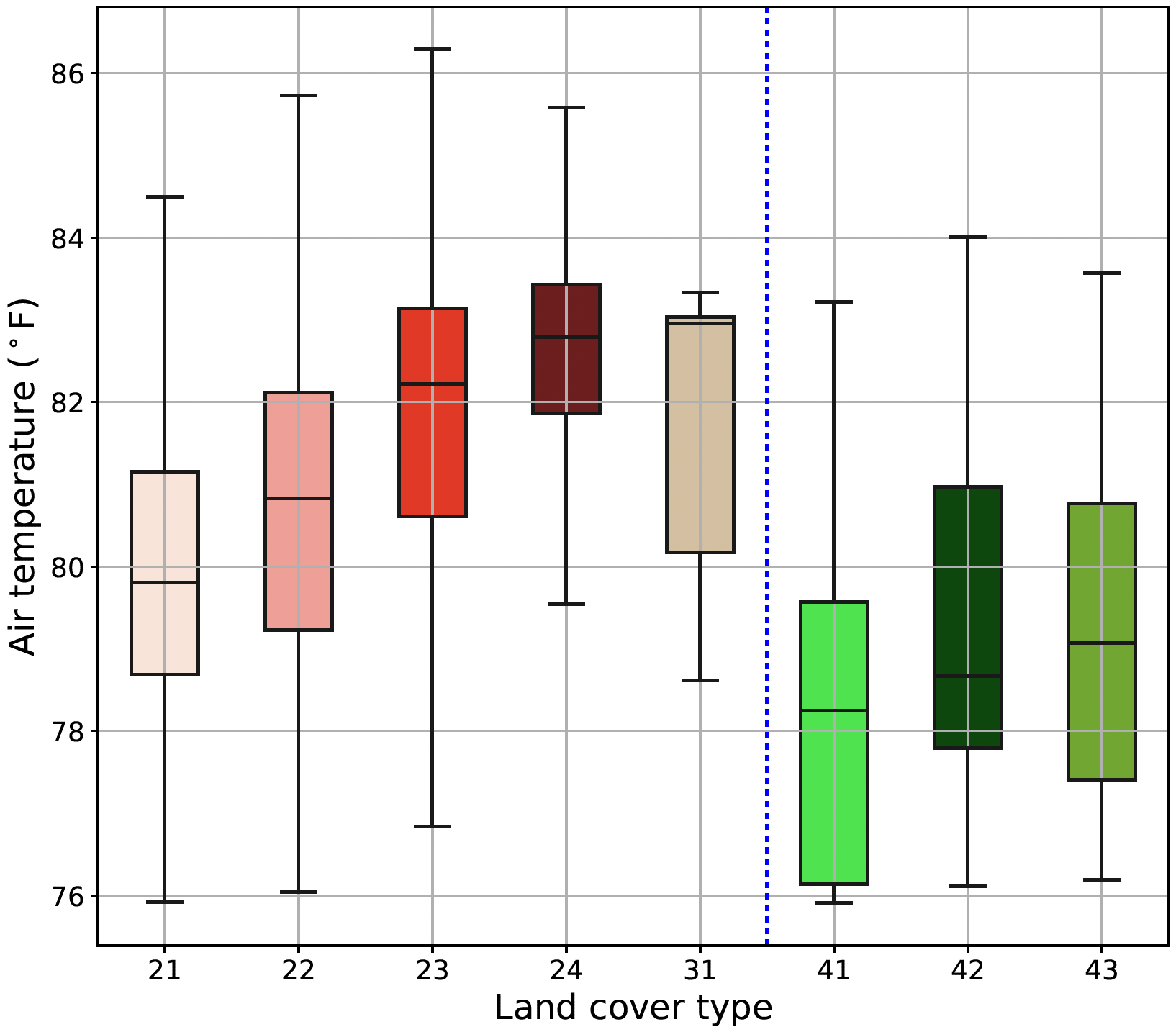}}
\caption{Distribution of the indirect effect on temperature from NDVI for both linear and NN-based models, where the bar shows the $25^{th}$, $50^{th}$ (i.e., median), and $75^{th}$ percentiles while the whiskers extend to show the rest of the distribution. Bars to the left of the blue dashed line represent land cover types with high building intensity, while those to the right represent land cover types with low building intensity.}
\label{fig:3}
\vspace{-3mm}
\end{figure*}

\begin{wraptable}{R}{0.6\textwidth}
\vspace{-4mm}
\caption{Prediction statistics for temperature using linear and NN-based models in low, medium, and high-NDVI regions}
\label{tab:3}
\centering
\setlength{\tabcolsep}{0.5em}
\begin{tabular}{lc|cccccc} 
\hline
\multirow{2}{*}{} & \multirow{2}{*}{$U_{\phi}$} & \multicolumn{2}{c}{Low NDVI} & \multicolumn{2}{c}{Medium NDVI} & \multicolumn{2}{c}{High NDVI}  \\
                  &                    & $R^2 \uparrow$ & MAE $\downarrow$                & $R^2 \uparrow$ & MAE $\downarrow$                    & $R^2 \uparrow$ & MAE $\downarrow$                   \\ 
\hline
Linear            &                    & 0.358  & 1.100               & 0.356  & 1.234                  & 0.230  & 1.321                 \\
Linear            & \checkmark             & 0.386  & 1.053               & 0.387  & 1.176                  & 0.243  & 1.289                 \\ 
\hline
CNN         &                    & 0.599  & 0.812               & 0.628  & 0.889                  & 0.472  & 1.011                 \\
CNN         & \checkmark             & 0.584  & 0.809               & 0.609  & 0.892                  & 0.430  & 1.056                 \\
U-Net       &                    & 0.404  & 1.051               & 0.434  & 1.127                  & 0.326  & 1.193                 \\
U-Net       & \checkmark             & 0.415  & 1.036               & 0.450  & 1.108                  & 0.341  & 1.168                 \\
\hline
\end{tabular}
\vspace{-1mm}
\end{wraptable}

To further demonstrate the utility of our method, we conduct a case study on a real-world dataset using the actual temperature observations instead of the synthetically generated temperature data. In this study, we use both NDVI and albedo as our treatments, denoted as $T_1$ and $T_2$, respectively, and calculate direct, indirect, and total effects using \eqref{eq:3}-\eqref{eq:5} as we only have the observed values of temperature instead of the full dose-response functions. All other experimental settings (i.e., NN architecture, optimization scheme, learning rate, etc.) remain consistent with those used in the semi-synthetic experiment.

We estimate the direct, indirect, and total effects from NDVI and compare them with the ground truth values of temperature as shown in Figure \ref{fig:3}. These results qualitatively show that our method is able to infer indirect effects that better reflect the expected behavior of the treatments we consider. For example, we expect that the magnitude of indirect effect from vegetation (i.e., NDVI) should decrease as building intensity increases. There are at least two reasons to expect this behavior: (1) we expect less horizontal mixing in areas with higher surface roughness (i.e., greater building density); and (2) high urban NDVI values likely indicate lower-height vegetation and grasses, whereas high suburban/rural NDVI values more likely indicate mature, taller trees. From Figure \ref{fig:3}, we observe that both the CNN-based (Figure \ref{fig:3b}) and U-Net-based (Figure \ref{fig:3c}) models predict a larger magnitude of indirect effect from NDVI in land cover types with low building intensity (classes 21, 22, 23, 24, and 31) compared to those with high building intensity (classes 41, 42, and 43), which is consistent with our expectations and the ground truth temperature (Figure \ref{fig:3d}). Practically, these results suggest vegetation decreases urban temperatures, but to a lesser extent than vegetation in suburban neighborhoods. In contrast, this pattern is less obvious in the predictions from the linear model (Figure \ref{fig:3a}).

Additionally, we try directly predicting the temperatures using the spatial regression models in regions with low (less than $30^{th}$ percentile), medium (between $30^{th}$ and $70^{th}$ percentile), and high (greater than $70^{th}$ percentile) NDVI. As shown in Table \ref{tab:3}, the NN-based models demonstrate superior overall performance in temperature prediction.

\section{Limitations}
\label{sec:7}

One limitation of the proposed lies in the model form of \eqref{eq:9}. A more flexible model could account for non-linear direct effects, as well as the interaction between direct and indirect effects, though incorporating additional non-linearities might compromise model identifiability. Additionally, our experiments have so far involved only environmental spatial data. Further evaluation is necessary to extend the application of our method to other scientific domains.

\section{Discussion and Conclusion}
\label{sec:8}

Many phenomena in the real-world exhibit non-linear or non-stationary interference, thus the development of methods to deal with this case are required to estimate causal effects. To this end, we present an NN-based framework with an approximate GP for causal inference with spatial interference and unobserved confounder. With quantitative evaluation on synthetic and semi-synthetic data, along with qualitative case study on real-world data, we show that the proposed method offers a solution to capture these nonlinear relationships, which allows for both better estimates of causal effects and more accurate predictions on the target variables.

\bibliographystyle{unsrtnat}
\bibliography{references}
\newpage

\appendix
\renewcommand\thesection{\Alph{section}}
\setcounter{section}{0}

\begin{table}[t!]
\begin{minipage}[c]{0.37\textwidth}
\centering
\caption{Land cover classes corresponding to the color codes in Figure \ref{fig:2c}.}
\label{tab:4}
\centering
\begin{tabular}{c|c} 
\hline
Code & Land Cover Type               \\ 
\hline
11   & Open Water                    \\
21   & Developed, Open Space         \\
22   & Developed, Low Intensity      \\
23   & Developed, Medium Intensity   \\
24   & Developed, High Intensity     \\
31   & Barren Land                   \\
41   & Deciduous Forest              \\
42   & Evergreen Forest              \\
43   & Mixed Forest                  \\
52   & Shrub                         \\
71   & Grassland, Herbaceous         \\
81   & Pasture                       \\
82   & Cultivated Crops              \\
90   & Woody Wetlands                \\
95   & Emergent Herbaceous Wetlands  \\
\hline
\end{tabular}
\vspace{-2mm}
\end{minipage}\hfill
\begin{minipage}[c]{0.6\textwidth}
\centering
\caption{Details of neural network architectures used in the experiments on synthetic, semi-synthetic, and real-world data. Here $\sigma$, $l$, and $\epsilon$ denote the standard deviation, lengthscale, noise level of the radial basis function (RBF) kernel for $U_{\phi}$.}
\label{tab:5}
\centering
\setlength{\tabcolsep}{0.2em}
\begin{tabular}{ll|ccc} 
\hline
 &  & $f$ & $g$ & $U_{\phi}$ \\ 
\hline
\multicolumn{2}{l|}{Synthetic}                                                       
& \begin{tabular}[c]{@{}c@{}}MLP\\width = 256\\depth = 3\end{tabular}     & \begin{tabular}[c]{@{}c@{}}MLP\\width = 256\\depth = 3\end{tabular} & \begin{tabular}[c]{@{}c@{}}RBF kernel\\$\sigma = 1$\\$l = 0.5$\\$\epsilon = 0.5$\end{tabular}    \\
\multirow{2}{*}{\begin{tabular}[c]{@{}l@{}}Semi-\\synthetic\end{tabular}} & \begin{tabular}[c]{@{}l@{}}CNN-\\based\end{tabular}   & \begin{tabular}[c]{@{}c@{}}CNN\\\#channels = 64\\depth = 9\end{tabular} & \begin{tabular}[c]{@{}c@{}}MLP\\width = 128\\depth = 2\end{tabular} & \begin{tabular}[c]{@{}c@{}}RBF kernel\\$\sigma = 1$\\$l = 0.005$\\$\epsilon = 0.1$\end{tabular}  \\
& \begin{tabular}[c]{@{}l@{}}U-Net-\\based\end{tabular} & \begin{tabular}[c]{@{}c@{}}U-Net\\depth = 3\\padding = 1\end{tabular}   & \begin{tabular}[c]{@{}c@{}}MLP\\width = 128\\depth = 2\end{tabular} & \begin{tabular}[c]{@{}c@{}}RBF kernel\\$\sigma = 1$\\$l = 0.005$\\$\epsilon = 0.1$\end{tabular}  \\
\multirow{2}{*}{\begin{tabular}[c]{@{}l@{}}Real-\\world\end{tabular}}     & \begin{tabular}[c]{@{}l@{}}CNN-\\based\end{tabular}   & \begin{tabular}[c]{@{}c@{}}CNN\\\#channels = 64\\depth = 9\end{tabular} & \begin{tabular}[c]{@{}c@{}}MLP\\width = 128\\depth = 2\end{tabular} & \begin{tabular}[c]{@{}c@{}}RBF kernel\\$\sigma = 1$\\$l = 100$\\$\epsilon = 0.1$\end{tabular}    \\
& \begin{tabular}[c]{@{}l@{}}U-Net-\\based\end{tabular} & \begin{tabular}[c]{@{}c@{}}CNN\\\#channels = 64\\depth = 9\end{tabular} & \begin{tabular}[c]{@{}c@{}}MLP\\width = 128\\depth = 2\end{tabular} & \begin{tabular}[c]{@{}c@{}}RBF kernel\\$\sigma = 1$\\$l = 100$\\$\epsilon = 0.1$\end{tabular}    \\
\hline
\end{tabular}
\end{minipage}
\end{table}

\section{Experimental Details on Geospatial Data}
\label{appx:A}

\subsection{Feature Details of Geospatial Dataset}
\label{appx:A.1}
As depicted in Figure \ref{fig:2}, the geospatial dataset consists of 4 features: the surface air temperature, the albedo values (using Sentinel-2), the normalized difference vegetation index (NDVI, calculated using Sentinel-2), and the National Land Cover Database (NLCD). The details of data collection are specified below and can also be found in \citet{calhoun2024estimating}.

\subsubsection{Surface Air Temperature}
\label{appx:A.1.1}
The temperature data is collected through the National Oceanic and Atmospheric Administration (NOAA) urban heat island mapping campaign using the approach described in \citet{shandas2019integrating}. When collecting the data, volunteers attached temperature sensors to their vehicles and drove around a traversal path at three periods of time (morning, afternoon, and evening) throughout the day. We use the data collected during the evening since this is the time when the temperature differences were greatest. The dataset is collected as point-referenced data at higher than 10 meter resolution, and we use the GDAL \cite{gdal} utility to convert the point data to a rasterized format. The temperature values are averaged if there exist more than one points in each 10-meter by 10-meter pixel. This data conversion resulted in a 2253-by-2307 pixel image of the data, with 12,448 temperature measurements, and the following albedo values, vegetation indices, and land cover types are subset to match these dimensions.

\subsubsection{Albedo}
\label{appx:A.1.2}
The albedo is calculated using the method defined in \citet{bonafoni2020albedo} based on Sentinel-2 surface reflectance data. The Sentinel-2 data is collected using Google Earth using the least cloudy data within 2 months of the date of the temperature collection.

\begin{figure*}[t!]
\centering
\begin{tikzpicture}
\node (img) {\includegraphics[scale=0.35]{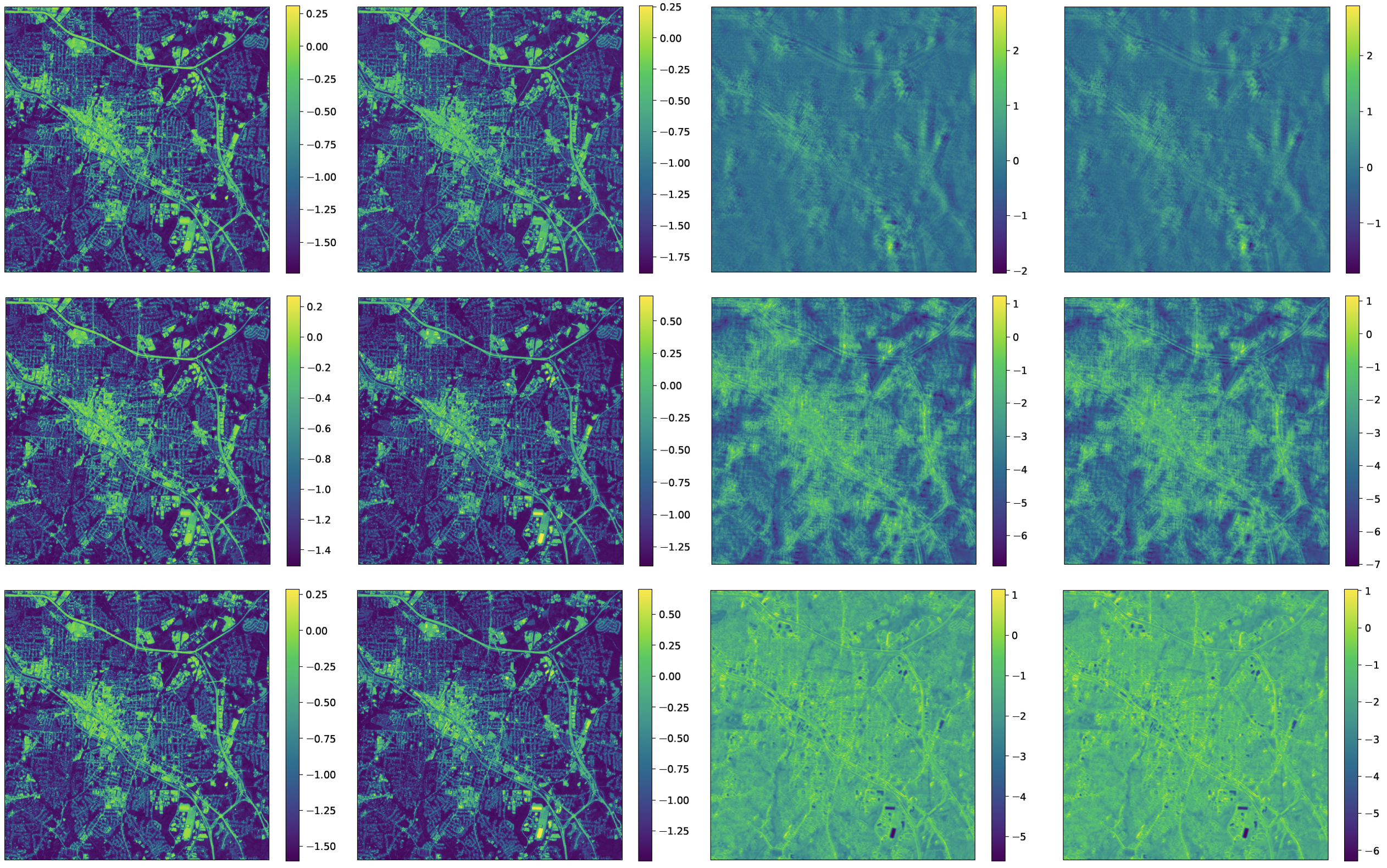}};
\node[below=of img, node distance=0cm, xshift=-6.2cm, yshift=1cm, text width=2cm, align=center] {\small Direct effect from NDVI};
\node[below=of img, node distance=0cm, xshift=-2.4cm, yshift=1cm, text width=2cm, align=center] {\small Direct effect from albedo};
\node[below=of img, node distance=0cm, xshift=1.9cm, yshift=1cm, text width=2cm, align=center] {\small Indirect effect from NDVI};
\node[below=of img, node distance=0cm, xshift=5.8cm, yshift=1cm, text width=2cm, align=center] {\small Indirect effect from albedo};
\node[left=of img, node distance=0cm, rotate=90, anchor=center, yshift=-0.6cm, xshift=3.3cm, text width=2cm, align=center] {\small Linear model};
\node[left=of img, node distance=0cm, rotate=90, anchor=center, yshift=-0.6cm, xshift=0cm, text width=2cm, align=center] {\small $f_{\theta_t}$: CNN, $g_{\theta_x}$: MLP};
\node[left=of img, node distance=0cm, rotate=90, anchor=center, yshift=-0.6cm, xshift=-3.2cm, text width=2cm, align=center] {\small $f_{\theta_t}$: U-Net, $g_{\theta_x}$: MLP};
\end{tikzpicture}
\caption{Visualization of the direct and indirect effects from NDVI and albedo on the surface air temperature in the downtown region of Durham.}
\label{fig:4}
\end{figure*}
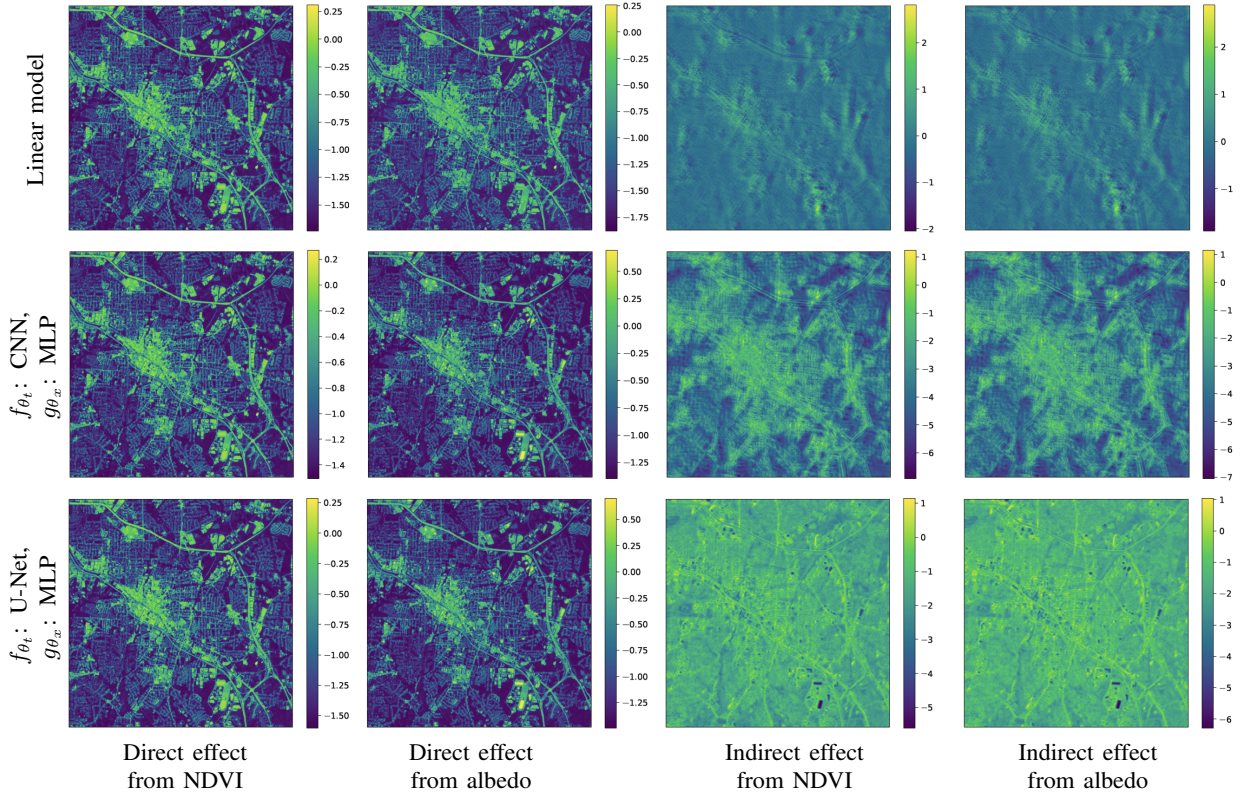

\subsubsection{Normalized Difference Vegetation Index (NDVI)}
\label{appx:A.1.3}
The NDVI is calculated based on the same Sentinel-2 data used for albedo calculation. Specifically, NDVI is calculated using the near-infrared (NIR) and red bands (R) of a satellite image as shown below:
\begin{equation}
\text{NDVI} = \frac{\text{NIR} - \text{R}}{\text{NIR} + \text{R}}.
\end{equation}

\subsubsection{National Land Cover Database (NLCD)}
\label{appx:A.1.4}
The NLCD was collected in 2021, the same year the temperature data was collected. We then upsample the NLCD from its original resolution
of 30 meters to match the albedo and NDVI resolution at 10 meters. There are 15 classes color coded in Figure \ref{fig:2c}, which correspond to 15 categories of land cover as shown in Table \ref{tab:4}.

\subsection{Computing Resources}
\label{appx:A.2}
The code implementations of the proposed method are developed using the PyTorch library. The experiment on the synthetic dataset is relatively small in scale and can be executed locally. The experiments on the semi-synthetic and real-world geospatial datasets are conducted on a computer cluster equipped with NVIDIA A6000 GPU nodes.

\section{Further Details of Model Architectures}
\label{appx:B}

We present the details of the model architectures, including the width and depth of the neural networks and the type and parameters of the kernel function used for the approximate GP, in Table \ref{tab:5}. It is worth noting that a significantly larger lengthscale is used for the real-world geospatial dataset compared to the semi-synthetic dataset, as the spatial coordinates in the real-world dataset are not scaled to the range $0 < \boldsymbol{s} < 1$.

\section{Visualizations of Direct, Indirect, and Total Effects}
\label{appx:C}
For the real-world experiment on geospatial data, in addition to visualizing the indirect effect, we also plot the spatial distributions of the direct and indirect effects from NDVI and albedo on the surface air temperature for the entire downtown region of Durham, NC as shown in Figure \ref{fig:4}. The results indicate that both linear and nonlinear spatial regression models yield similar predictions for the direct effects, which is expected given that $\alpha_m T_{i,m}$ is modeled as a linear term in \eqref{eq:9}. However, for the indirect effects, nonlinear models, particularly the CNN-based model (second row in Figure \ref{fig:4}), provide more \emph{fine-grained} predictions in the central area of downtown Durham. Specifically, the CNN-based model predicts close-to-zero or positive indirect effects in high-intensity regions and negative indirect effects in low-intensity regions. This suggests that higher vegetation index and albedo in low-intensity regions contribute to a \emph{decrease} in surface air temperature, while the opposite is true for high-intensity regions, aligning with our expectations.

\end{document}